\documentclass{egpubl}
\JournalSubmission    
\usepackage[T1]{fontenc}
\usepackage{dfadobe}  
\usepackage{cite}  
\BibtexOrBiblatex
\electronicVersion
\PrintedOrElectronic

\ifpdf \usepackage[pdftex]{graphicx} \pdfcompresslevel=9
\else \usepackage[dvips]{graphicx} \fi

\usepackage{egweblnk} 
\usepackage{hyperref}
\usepackage{etoolbox,lipsum}
\usepackage{subfigure}
\usepackage{color}

\usepackage{amssymb} 
\usepackage{threeparttable} 
\usepackage{booktabs} 
\usepackage{float} 

\newcommand{\added}[1]{\textcolor{blue}{#1}}

\title[SVBRDF Recovery From a Single Image with Highlights using 
a Pretrained GAN]
      {SVBRDF Recovery From a Single Image with Highlights \\ using a Pretrained Generative Adversarial Network}

	  \author[T. Wen, B. Wang, L. Zhang, J. Guo and N. Holzschuch]
{\parbox{\textwidth}{\centering Tao Wen$^1$, Beibei Wang\thanks{Corresponding author\\{Email address: beibei.wang@njust.edu.cn}}$^{1,2}$, Lei Zhang$^2$, Jie Guo$^3$ and Nicolas Holzschuch$^4$}
        \\
{\parbox{\textwidth}{\centering $^1$Nanjing University of Science and Technology\\
$^2$The Hong Kong Polytechnic University\\
$^3$Nanjing University\\
$^4$Univ. Grenoble Alpes, Inria, CNRS, Grenoble INP, LJK
       }
}
}

\begin{document}
\teaser{
 \includegraphics[width=1.0\linewidth]{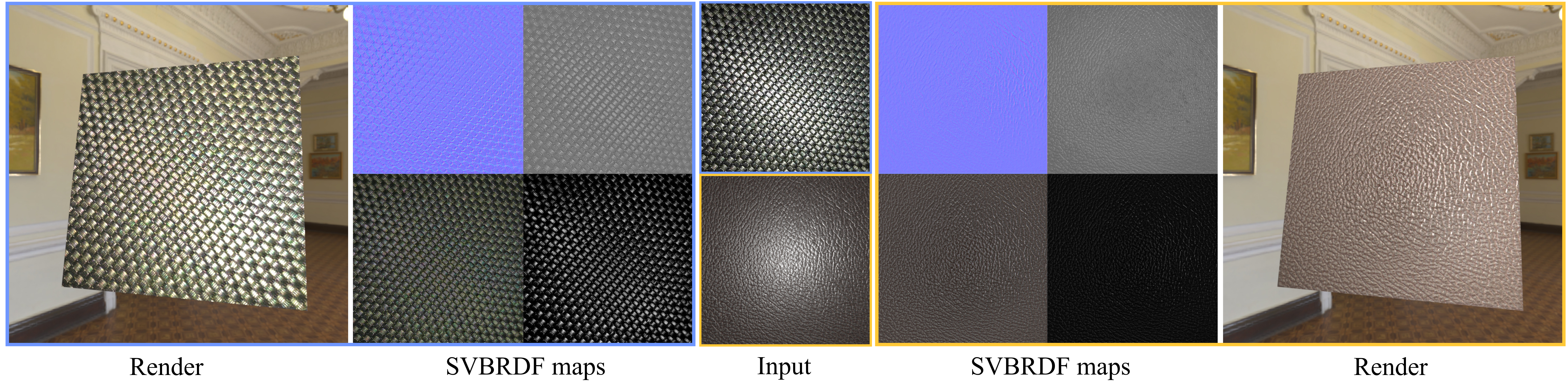}
 \centering
 \caption{Our method generates high-quality and natural SVBRDFs maps from a single input photograph with overexposed highlight regions, and provides vivid rendering. Our Fourier-based loss function separates the effects due to the specular highlight from those due to the material. }
\label{fig:teaser}
}

\maketitle
\begin{abstract}
Spatially-varying bi-directional reflectance distribution functions (SVBRDFs) are crucial for designers to incorporate new materials in virtual scenes, making them look more realistic. Reconstruction of SVBRDFs is a long-standing problem. Existing methods either rely on extensive acquisition system or require huge datasets which are nontrivial to acquire. We aim to recover SVBRDFs from a single image, without any datasets. A single image contains incomplete information about the SVBRDF, making the reconstruction task highly ill-posed. It is also difficult to separate between the changes in color that are caused by the material and those caused by the illumination, without the prior knowledge learned from the dataset. In this paper, we use an unsupervised generative adversarial neural network (GAN) to recover SVBRDFs maps with a single image as input. To better separate the effects due to illumination from the effects due to the material, we add the hypothesis that the material is stationary and introduce a new loss function based on Fourier coefficients to enforce this stationarity. For efficiency, we train the network in two stages: reusing a trained model to initialize the SVBRDFs and fine-tune it based on the input image. Our method  generates high-quality SVBRDFs maps from a single input photograph, and provides more vivid rendering results compared to previous work. The two-stage training boosts runtime performance, making it 8 times faster than previous work. 

\printccsdesc   
\end{abstract}  


\section{Introduction}

The reconstruction of real world material appearance is a long standing problem in computer graphics and vision. The reflectance parameters of opaque materials can be modeled by the 6D spatially-varying bi-directional reflectance distribution function (SVBRDF). It is difficult to recover the SVBRDFs of a real world material because of its high dimensionality and the inherent ambiguity of the unknown parameters: color variations could be caused by changes in any material parameter: albedo, roughness or normal. Several previous methods required complex acquisition equipments to densely sample materials in different light and view directions. These can faithfully capture the appearance parameters of a material, but are also very expensive and time-consuming, limiting the accessibility. 

Recent works have shown that it is possible to recover the SVBRDFs from a few photos, or even a single image, of the material~\cite{deschaintre2018single, li2018materials, gao2019deep, Guo2020MaterialGANRC, Guo2021HighlightawareTN}. These lightweight methods used deep neural networks to capture four SVBRDF maps (diffuse and specular albedo, normal map and roughness parameters) from photographs of a material. They usually rely on convolutional neural networks (CNNs), trained on  synthetic images and corresponding SVBRDF maps, to model the appearance of real world materials. 

These deep learning-based methods are {\em supervised} and require large training datasets. These datasets are difficult to acquire~\cite{li2017modeling, deschaintre2018single, aittala2015two}. Existing methods either need professional designers to generate procedural models, or rely on numerous samples of real world materials. A concurrent approach relies on generative adversarial network (GAN)~\cite{goodfellow2014generative} to avoid heavy work in dataset collection. Zhao et al.~\cite{Zhao2020JointSR} proposed the first approach to exploit GAN architecture for unsupervised SVBRDF maps recovery. They rely on a two-stream generator to train the SVBRDF maps (diffuse, specular, normal, roughness) and calculate the adversarial loss. Their network is able to predct plausible SVBRDF maps from a single input image, and does not require any dataset. They also provide high quality texture synthesis through a well-designed encoder-decoder structure. 

When the input image includes an intense specular highlight, it is difficult for acquisition methods to separate between albedo and illumination. As the specular highlight has a large intensity, it gets a strong priority in the learning process, often resulting in bright spots at the center of the albedo maps. To solve this issue, we need to introduce a specific constraint: we make the hypothesis that the material we acquire is stationary (its features repeat themselves after a certain period). We enforce stationarity in the reconstructed SVBRDF maps using a new loss function based on the Fourier transforms of the SVBRDF maps. Also, the rendering loss function based on pixel-wise comparisons do not work well when the exact positions of the camera and the light source are not well known. We introduce a new loss function, based on the perceptual difference between the input image and the reconstructed image~\cite{johnson2016perceptual}. Combined together, these new loss functions generate high-quality SVBRDF maps from a single input image. In particular, when there are overexposed highlights in the input photograph, our method generates more reasonable SVBRDF maps compare to previous works, leading to more realistic results when re-rendering with different viewing and illumination.

To speed-up the reconstruction process, we also introduce a two-stage training strategy: we pretrain our network on a single material, which provides the starting parameters for the training on each input image. This pretraining strategy makes the treatment of an image 8 times faster.

In summary, our contributions include:
\begin{itemize}
\item a Fourier loss function to enforce stationarity in the SVBRDFs, which produces more plausible results when the input image has intense highlights,
\item a perceptual loss function that measure the semantic similarity between the input image and re-rendering result, and 
\item a two-stage training strategy to speedup the train process without quality degradation.

\end{itemize}

The rest of the paper is organized as follows. In Sec.~\ref{sec:related}, we review previous works on SVBRDFs recovery. As our method builds on Zhao et al.\cite{Zhao2020JointSR}, we present this method in depth in  Sec.~\ref{sec:background}. We present our method and implementation in Sec.~\ref{sec:recover}. We show and discuss our results in Sec.~\ref{sec:results}, and conclude in Sec.~\ref{sec:conclusion}.

\section{Related work}
\label{sec:related}

The problem of appearance capture has been extensively researched. Please refer to Guarnera et al.~\cite{guarnera2016brdf} and Gao et al.~\cite{gao2019deep} for a more comprehensive introduction. In this paper, we focus on light-weight appearance capture, which can be grouped into multi-image methods and single-image methods according to the number of input images. 


\subsection{Multi-image appearance modeling} 

\textbf{Non-learning based methods.} With multiple images as input, previous works can capture the SVBRDF based on optimization usually with some domain-specific priors or assumptions, e.g. the known illumination~\cite{chandraker2014shape,hui2015dictionary,riviere2016mobile}, or sparsity in some domain ~\cite{hui2017reflectance,dong2014appearance,Xia:2016:Shape}. Aittala et al.~\cite{aittala2015two} used two photographs (one with flash and one without) to recover the reflectance, assuming the maps are stationary. Xu et al.~\cite{Xu2016NearField} used two images from a near-field perspective camera, and assume spatial relation for reflectance recovery.

\textbf{Learning based methods.} Recently, deep learning has been widely used in appearance modeling. Deschaintre et al.~\cite{Valentin2019Flexible} extracted the feature from each input image via a single-image appearance modeling network (similar to~\cite{deschaintre2018single}) and then fused the features for SVBRDF recovery, to support arbitrary number of input images. Gao et al.~\cite{gao2019deep} proposed an auto-encoder to extract the latent space from SVBRDFs as a material 'prior', and then optimized material maps in this latent space to better leverage the inherent connections between maps. However, their method needs an initial value of SVBRDF maps. Guo et al.~\cite{Guo2020MaterialGANRC} trained a MaterialGAN to produce plausible material maps from a small number (3-7) of images. They used three optimizing strategies for the intermediate vector and noise vector in the latent space to learn the correlations in SVBRDF parameters. In order to tackle the shape/SVBRDF ambiguity, Boss et al.~\cite{Boss2020-TwoShotShapeAndBrdf} designed a cascaded network for shape, illumination and SVBRDF estimation, using two images captured by a cellphone with flash both on and off.

\begin{figure*}[htb]
	\centering
	\includegraphics[width=0.993\linewidth]{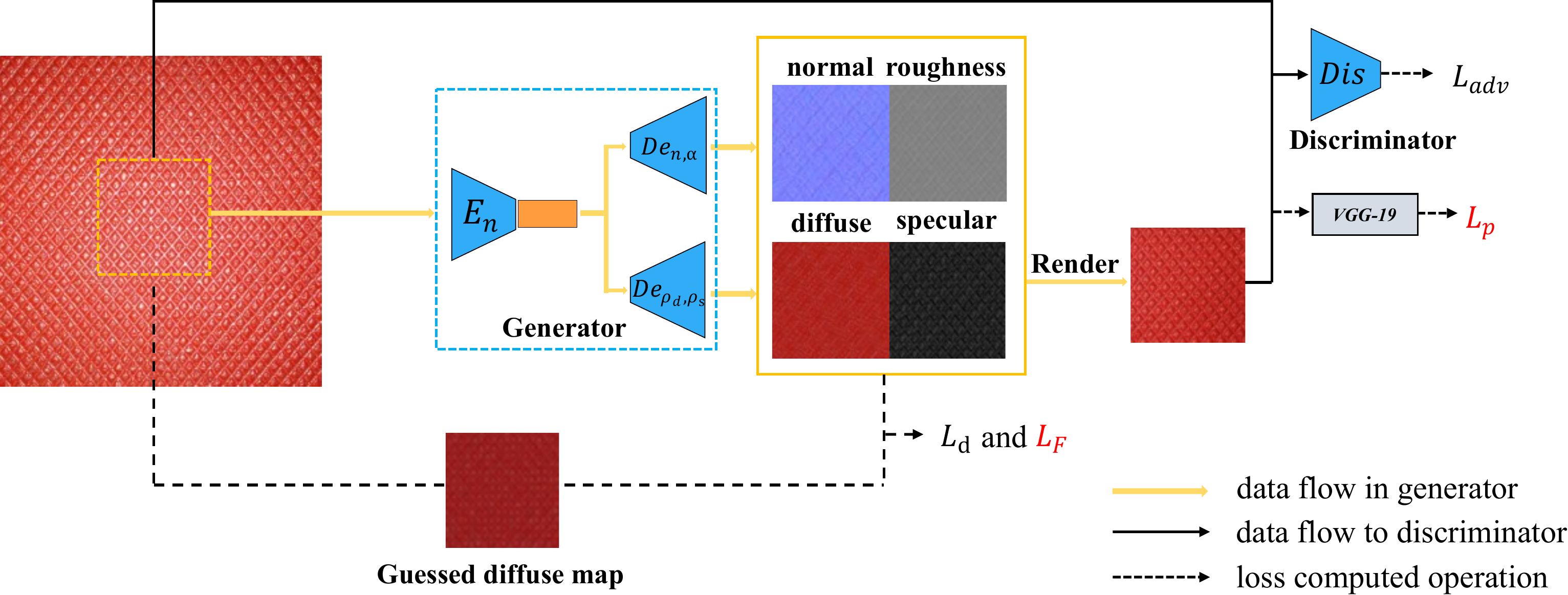}
	\caption{\label{fig:newframework}%
		 Our method is similar to Zhao et al.~\cite{Zhao2020JointSR}, with several key differences: we reduce the number of de-conv layers in the decoder to let the $Generator$ output SVBRDF maps of same resolution as input. The predicted maps are used to calculate the diffuse loss $L_{d}$ and Fourier loss $L_{F}$. The input tile and re-rendered image are feed into discriminator to get the adversial loss $L_{adv}$ and a pretrained VGG-19 network to get the perceptual loss $L_{p}$. Our new loss terms are shown in red in the architecture.
	}
\end{figure*}

\subsection{Single-image appearance modeling} 

Another group of works only use single image as input. Aittala et al.~\cite{aittala2016reflectance} proposed a convolutional neural network (CNN) to extract a neural Gram-matrix texture descriptor from a single image to estimate the reflectance properties of a stationary textured materials. Under the same assuption, Zhao et al.~\cite{Zhao2020JointSR} proposed an unsupervised generative adversarial network for joint SVBRDF recovery and synthesis. When the input image has intense highlights, their method confuses material properties and tends to produce maps with a bright spot for the specular albedo. It also takes a long time to process each input image. Our method addresses both issues. 

Li et al.~\cite{li2017modeling} trained a CNN with a novel self-augmentation training strategy, which requires only a small number of labeled SVBRDF training pairs, to learn a large number of unlabeled photos of spatially varying materials. Ye et al.~\cite{ye2018single} improved on this method and completely eliminated the need for labeled training dataset. Deschaintre et al.~\cite{deschaintre2018single} proposed a secondary network to extract global features from each stage of an U-net architecture. They also introduce a rendering loss to enhance the estimated reflectance parameters by comparing the appearance rendered of predicted SVBRDF maps with the input image. Li et al.~\cite{li2018materials} designed an in-network rendering layer to regress SVBRDF maps from single image and a material classifier to constrain the latent representation of a CNN. They also utilized a densely-connected conditional random fields module to further refine the results. A current work by Guo et al.~\cite{Guo2021HighlightawareTN} proposed a new convolution variant called highlight-aware convolution (HA-convolution). They train the HA-convolution to ``guess'' the saturated pixels (specular highlight area) by the unsaturated area surrounded, making the extracted features more uniform. Their work achieve state-of-the-art performance on single-image SVBRDF acquisition and can well handle images with intense highlights. Compared to all these works, our method avoids the large dataset and learns the maps individually, under the stationary assumption.  

To remove the limitation of planar materials, Li et al.~\cite{li2018learning} proposed a cascaded network architecture to recover shape and SVBRDF simultaneously from a single image. This method is further extended to handle complex indoor scenes~\cite{li2020inverse}. 
 
In terms of predicting procedural texture parameters, Hu et al.~\cite{hu2019novel} introduced a novel framework for inverse procedural texture modeling: they trained an unsupervised clustering model to select a most appropriate procedural model and then used a CNN pool to map images to material parameters. 

\begin{figure*}[htbp]
	\centering
	\includegraphics[width=0.993\linewidth]{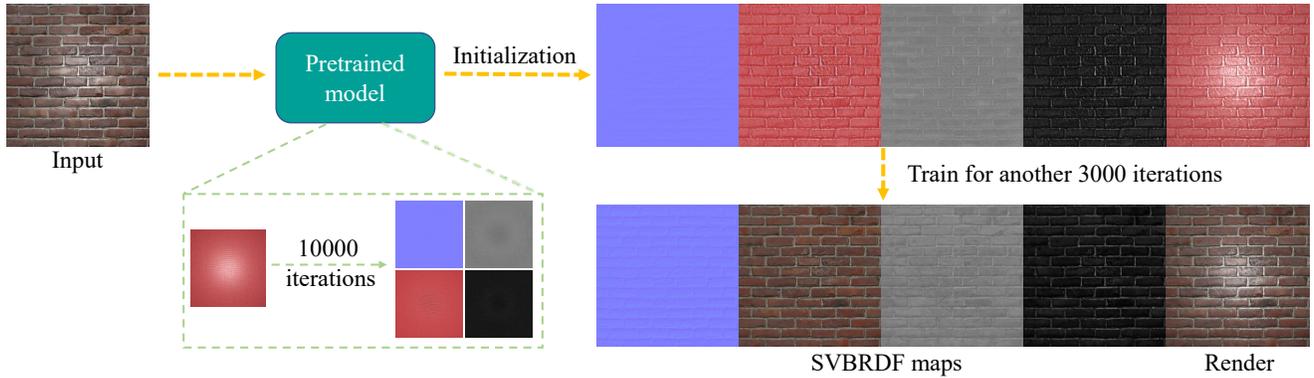}
	\caption{\label{fig:pretrain}%
		Our two-stage training strategy. At the first stage, we train the network on one image to get the pretrained model. For any new input images, we use this pretrained model as an initialization to start our second stage training. Note that at second stage, texture information already exists in the initialized SVBRDF maps, without extra training. 
	}
\end{figure*}

\begin{figure*}[htbp]
	\centering
	\includegraphics[width=0.9\linewidth]{synthetic-book.pdf}
	\caption{\label{fig:synthetic-book}%
		SVBRDF maps recovered from synthetic images of $1024 \times 1024$, compared with Gao et al.~\cite{gao2019deep}, Guo et al.~\cite{Guo2020MaterialGANRC}, Guo et al.~\cite{Guo2021HighlightawareTN} and Zhao et al.~\cite{Zhao2020JointSR}. 
		Note how the specular highlight at the center of the image is challenging for all acquisition methods, resulting in dark or bright areas in the specular albedo map, and flattened areas in the normal map. We use the network of~\cite{deschaintre2018single} as initialization for ~\cite{gao2019deep} and set the number of input images to one. Guo et al.~\cite{Guo2020MaterialGANRC} can only produce $256 \times 256$ maps so we scale them to proper size for comparison.
	}
\end{figure*}


\section{Background and motivation}
\label{sec:background}

In this section, we review the network by Zhao et al.~ \cite{Zhao2020JointSR}, since our method relies on it. Then, we discuss the motivation of our method.
\subsection{Background}

We assume the lighting and viewing directions to be identical, as it happens when the flash light is close to the camera lens. 
We generate four maps to represent the SVBRDF: the diffuse albedo $\rho_{d} \in \mathbb{R}^{3}$, specular albedo $\rho_{s} \in \mathbb{R}$, roughness $\alpha \in \mathbb{R}$, and surface normal $\mathbf{n} \in \mathbb{R}^{3}$. Our goal is to compute  $\mathbf{u} = {[\rho_{d},\rho_{s},\alpha,\mathbf{n}]}$. We use the Cook-Torrance reflectance model \cite{cook1982reflectance} for rendering.

\textbf{Network architecture.}
Zhao et al.~\cite{Zhao2020JointSR} proposed a generative adversarial network for SVBRDF recovery and synthesis. The network is unsupervised, thus no dataset is required for training. With a stationary image as an input, the network produces four SVBRDF maps, by training on different cropped tiles from the input image. The network consists of a \emph{two-stream generator} and a \emph{patch discriminator}. 

The generator includes an encoder and two decoders, where two groups of maps of the tiles are generated separately: normal and roughness, diffuse and specular. The maps are then used to render an image. Both this rendered image and the tile from the input image are fed to the discriminator to determine the correctness of the generated SVBRDF maps. Regarding the network structures, please see Zhao et al.~\cite{Zhao2020JointSR} for more details.

\textbf{Loss function.}
The loss function in Zhao et al.~\cite{Zhao2020JointSR} consists of a guessed diffuse map loss and the adversarial loss:
\begin{equation}\label{Lzhao}
\mathcal{L}_\mathrm{\added{Zhao}} = \lambda \mathcal{L}_\mathrm{GAN}(G,D) + \mathcal{L}_{d}(G),
\end{equation}
\begin{equation}\label{ganlossu}
\mathcal{L}_\mathrm{GAN}(G,D) = \mathbb{E}[\log {Dis}(\mathbf{x})] + \mathbb{E}[\log (1-{Dis}(\mathbf{y})],
\end{equation}
\begin{equation}\label{initd}
\mathcal{L}_{d}(G) = \mathbb{E}\left[\|\tilde{\mathbf{\rho_{d}}}-\mathbf{\rho_{d}}\|_{1}\right].
\end{equation}
The guessed diffuse map $\tilde{\mathbf{\rho_{d}}}$ is obtained via normalizing the input image, and considering the statistical distribution, which is used as the ground truth of the diffuse map, since the ground truth maps are absent. 

\subsection{Analysis}
When highlights exist in the input image, Zhao et al.~\cite{Zhao2020JointSR} fail to recover satisfactory SVBRDF maps due to the ambiguous highlight spot. The high intensity region is often classified as part of the specular albedo, resulting in wrong results for the roughness and specular maps (see Figure~\ref{fig:synthetic-brick}). This is an issue for all existing SVBRDF acquisition methods, due to the ambiguity between illumination and material. 

In this paper, we introduce extra information to resolve the ambiguity: the material maps we wish to recover are stationary, that is they repeat themselves after a certain period. As a consequence, the recovered maps should also be stationary. We enforce the stationarity of the recovered maps with a new loss function, based on their Fourier transform. We focus on the acquisition part of Zhao et al.~\cite{Zhao2020JointSR} and ignore the texture synthesis part, although it can be easily included.

Zhao et al.\cite{Zhao2020JointSR} also has an issue with computation time: as the network is trained from scratch on each individual image, processing can take up to 4 hours for a single image. We solve this issue with a  two-stage training strategy, using a pretrained model for initialization and a fine-tuning stage. The computation time is down to 30~mn for each image.


\section{Our method}
\label{sec:recover}

In this section, we propose a novel loss for the SVBRDF GAN~\cite{Zhao2020JointSR} to enforce the stationarity of SVBRDF maps and relax the pixel-wise connection between the guessed diffuse map and the input image. Then we present a two-stage training strategy to reduce the training time cost. Lastly, we show the implementation details.


\subsection{Stationarity-aware loss function}
\label{sec:loss}

We propose a joint loss, including a Fourier loss and a perceptual loss, where Fourier loss enforces the stationarity in SVBRDF maps and perceptual loss makes the rerendering result more plausible.
In Figure~\ref{fig:newframework}, we show the difference between our method and Zhao et al.~\cite{Zhao2020JointSR}.

\textbf{Fourier loss.} With a single image, it is difficult to separate between the color changes due to the material and those due to illumination. Without guidance, the network tends to place the highlights as part of the albedo or normal map. We introduce an extra constraint: the material should be stationary. In a stationary texture, variations are high-frequency and illumination effects are low-frequency. We introduce a new loss function based on Fourier analysis to enforce this stationarity: we compute the Fourier transform of the guessed diffuse map $\tilde{\mathbf{\rho_{d}}}$ and of the predicted SVBRDF maps $\mathbf{u}$(${[\rho_{d},\rho_{s},\alpha,\mathbf{n}]}$). We compute the Fourier loss function as the  $L_1$ loss in the logarithmic domain:
	\begin{equation}\label{Lf}
		\mathcal{L}_{F}(\mathbf{u},\tilde{\rho_{d}}) = \log\mathbb{E}\left[\| \mathrm{FFT}(\mathbf{u})-\mathrm{FFT}(\tilde{\rho_{d}})\|_{1}\right].
	\end{equation}
We use the fact that, after normalization, the guessed diffuse map has a stationary distribution of gray scale. 
With Fourier loss as a guidance on the frequency domain, the predicted SVBRDF maps will be less affected by the illumination and have similar variations as guessed diffuse map.


\textbf{Perceptual loss.} The exact lighting and viewing directions are sometimes unknown, especially for captured photographs. Using loss functions based on pixel-wise difference between input images and rendered images produces poor results, because we cannot guarantee the consistency of our rendering parameters: the rendered image is not rendered with exactly the same viewing and lighting condictions as the input image. To solve this issue, we use a perceptual loss function, to measure the semantic similarity between the input $I$ and the re-rendered result $R$, via a pretrained VGG-19 network:
	\begin{equation}\label{Lt}
		\mathcal{L}_{P}(I,R) = \mathbb{E}\left[\| \mathrm{VGG}(I)-\mathrm{VGG}(R)\|_{1}\right].
	\end{equation}
With this perceptual rendering loss, the re-rendering result of predicted SVBRDF maps is more realistic and reliable.

\textbf{Summary.} We present a joint loss function combining these three losses:
\begin{equation}\label{Lfinal}
\mathcal{L}_\mathrm{final} = \mathcal{L}_\mathrm{Zhao} + \lambda_{1}\mathcal{L}_{F}(\mathbf{u},\tilde{\rho_{d}}) +   \lambda_{2}\mathcal{L}_{P}(I,R)
\end{equation}
Trained with this joint loss function, the network can achieve better recovery result of SVBRDFs, leading to more reasonable rendering results with novel viewing/lighting directions, especially when handling images with intense highlights. We will show our recovery results in Sec.~\ref{sec:recovresult}.

\textbf{Discussion:} our loss function assumes that all maps being computed for the current SVBRDF have the similarly structured patterns with the same frequency, and that non-stationarity comes from illumination. This is a reasonable expectation for a large class of materials (leather, fabrics, wallpapers), but can be wrong for other materials, with local patterns. We also assume that all maps have the same frequency in their patterns, and that we can use the guessed diffuse map as a guide for learning in the other maps.


\begin{figure*}[htbp]
	\centering
	\includegraphics[width=0.9\linewidth]{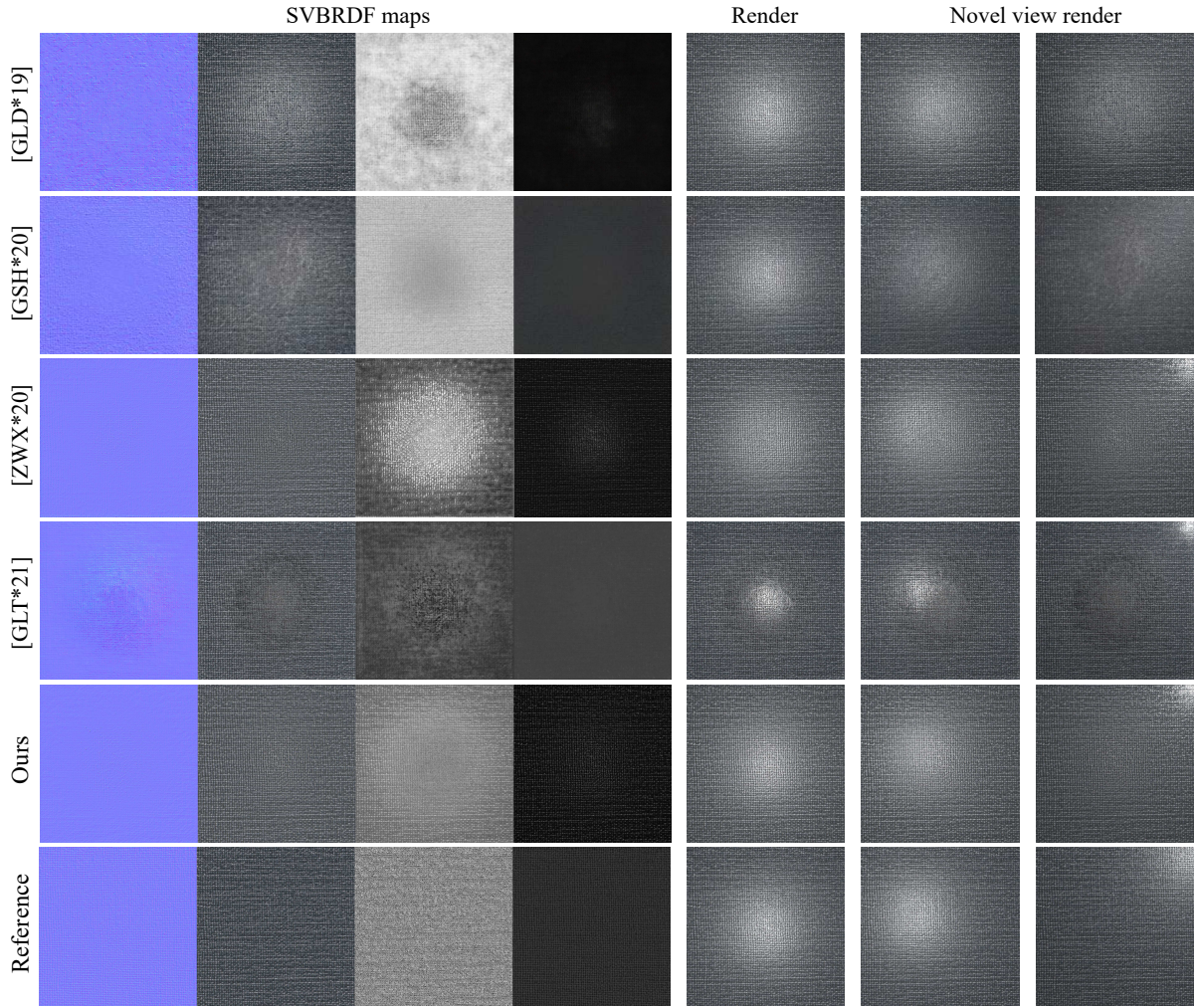}
	\caption{\label{fig:synthetic-brick}%
	     SVBRDF maps recovered from synthetic images of $1024 \times 1024$, compared with Gao et al.~\cite{gao2019deep}, Guo et al.~\cite{Guo2020MaterialGANRC}, Guo et al.~\cite{Guo2021HighlightawareTN} and Zhao et al.~\cite{Zhao2020JointSR}. Note how the specular highlight at the center of the image is challenging for all acquisition methods, resulting in dark or bright areas in the specular albedo map, and flattened areas in the normal map. We use the network of~\cite{deschaintre2018single} as initialization for ~\cite{gao2019deep} and set the number of input images to one. Guo et al.~\cite{Guo2020MaterialGANRC} can only produce $256 \times 256$ maps so we scale them to proper size for comparison.
 }
\end{figure*}
\subsection{Two-stage training strategy}

Training the network from scratch for each input image is time consuming: all parameters in the network have to be initialized as random value and trained over and over again for each new input. It can take several hours for the network to converge.

To improve this process, we propose a two-stage training strategy. We first train our network on an image (e.g. red book) for 10,000 steps to get a pretrained model. For a new input (e.g. brick), we use this pretrained model as an initialization of network parameters, and then train on this model for another 3,000 iterations to get the ``plausible'' SVBRDF maps. The key insight is that the generator acts as a prior knowledge about how the four maps look like in general after training for 10,000 steps:
\begin{itemize}
    \item the RGB value of normal map is close to light blue, due to the planar property of material,
    \item  the roughness map looks ``grey'',
    \item the specular map is ``dark'', and
    \item  the color of diffuse map mostly depends on the color of input image.
\end{itemize}
 Besides, as shown in Figure~\ref{fig:pretrain}, without any extra training, the generator could recover texture information of the new input image in SVBRDF maps (although the color is not ``correct''). Apparently, with the pretrained parameters as a good initialization, it become relatively easier for the network to get a plausible recovery result of SVBRDF maps, comparing to training from scratch.

We have tried different pretrained models with different input image and find little difference in the final results. We provide some results in Figure~\ref{fig:diffmodel}.

 \begin{figure*}[htbp]
	\centering
	\includegraphics[width=1\linewidth]{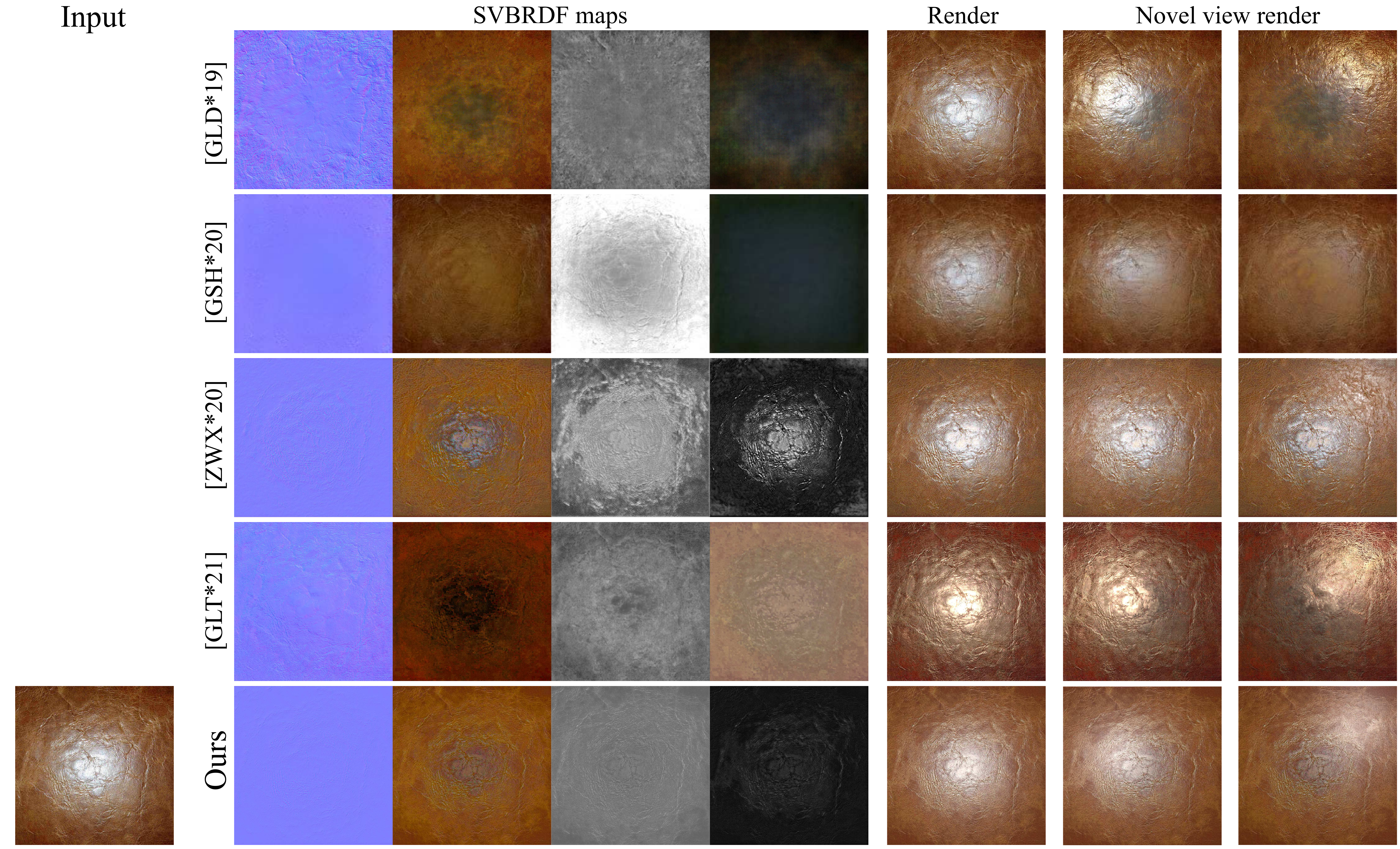}
	\caption{\label{fig:photo-leather}%
     SVBRDF maps recovered from real photos of $1024 \times 1024$, compared with Gao et al.~\cite{gao2019deep}, Guo et al.~\cite{Guo2020MaterialGANRC}, Guo et al.~\cite{Guo2021HighlightawareTN} and Zhao et al.~\cite{Zhao2020JointSR}. Note how the specular highlight at the center of the image is challenging for all acquisition methods. Our method produces more stationary SVBRDF maps and more plausible rendering results, compared to previous works.
 }
\end{figure*}
\subsection{Training and implementation}

We implemented our framework in TensorFlow. The generator and discriminator were trained using Adam optimizers with a fixed learning rate of 2e-5. We set the hyper-parameter $\lambda$, $\lambda_{1}$, $\lambda_{2}$ to 0.1, 0.1, 0.2 respectively.
At the first stage, we train our network on an arbitrary input image for 10000 iterations from scratch to obtain a pretrained model. Then with a new image as input, we fine-tune the pretrained network for another 3000 iterations to get a plausible result. It takes about 2 hours to get the pretrained model, which we then use for any new input images. It then costs about 30 minutes to train the model on a new image, using a RTX 2080Ti GPU. Note that training from scratch for a new input image, rather than using the pretrained model, requires 20,000 iterations with 4 hours, using the same GPU. Hence, the pretrained model represents an approximate 8 times speedup.


\section{Results and discussion}
\label{sec:results}

\begin{figure*}[htbp]
	\centering
	\includegraphics[width=1\linewidth]{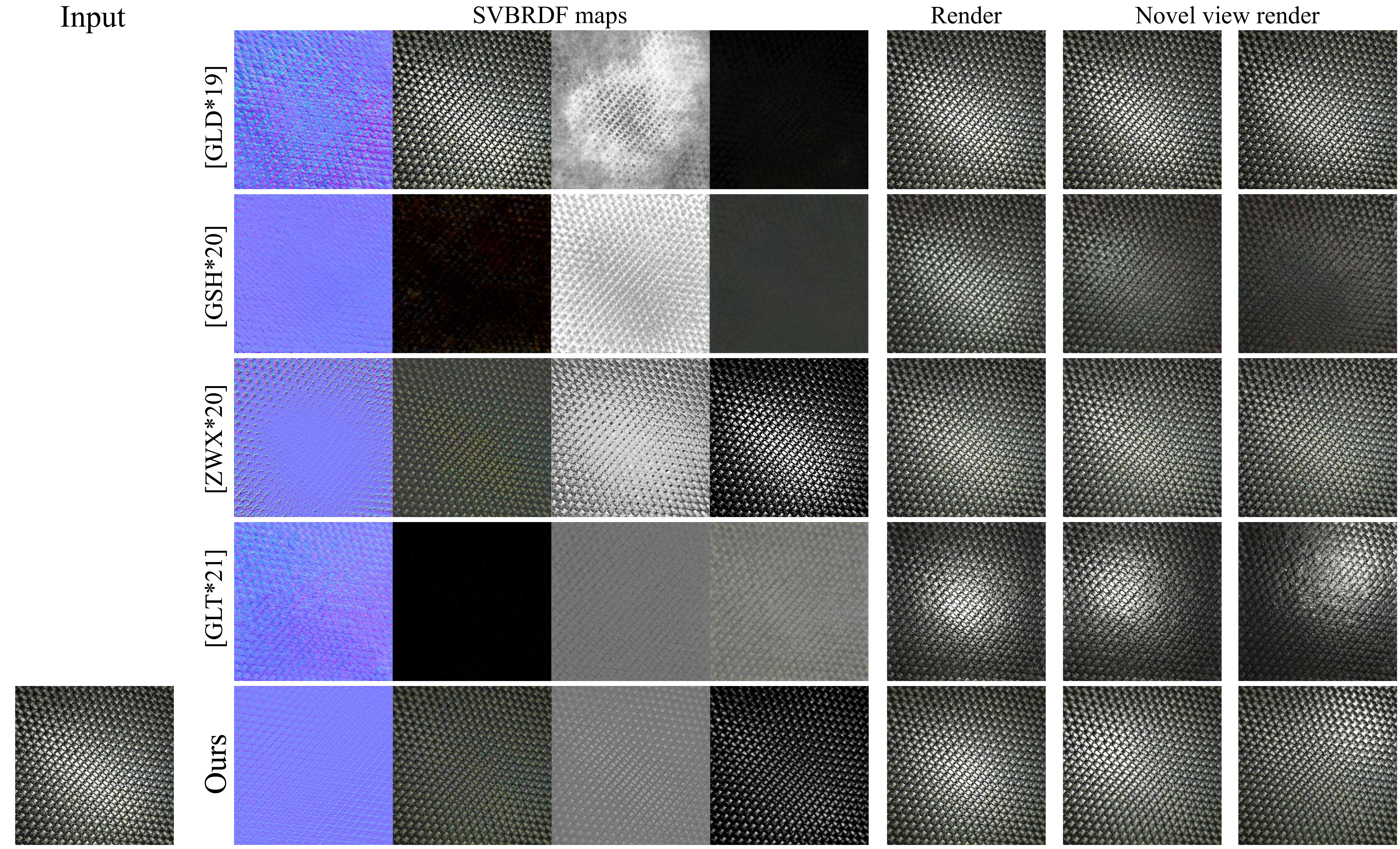}
	\caption{\label{fig:photo-plastic}%
	SVBRDF maps recovered from real photos of $1024 \times 1024$, compared with Gao et al.~\cite{gao2019deep}, Guo et al.~\cite{Guo2020MaterialGANRC}, Guo et al.~\cite{Guo2021HighlightawareTN} and Zhao et al.~\cite{Zhao2020JointSR}. Note how the specular highlight at the center of the image is challenging for all acquisition methods. Our method produces more stationary SVBRDF maps and more plausible rendering results, comparing to previous works.
 }
\end{figure*}

\begin{figure*}[htbp]
	\centering
	\includegraphics[width=0.95\linewidth]{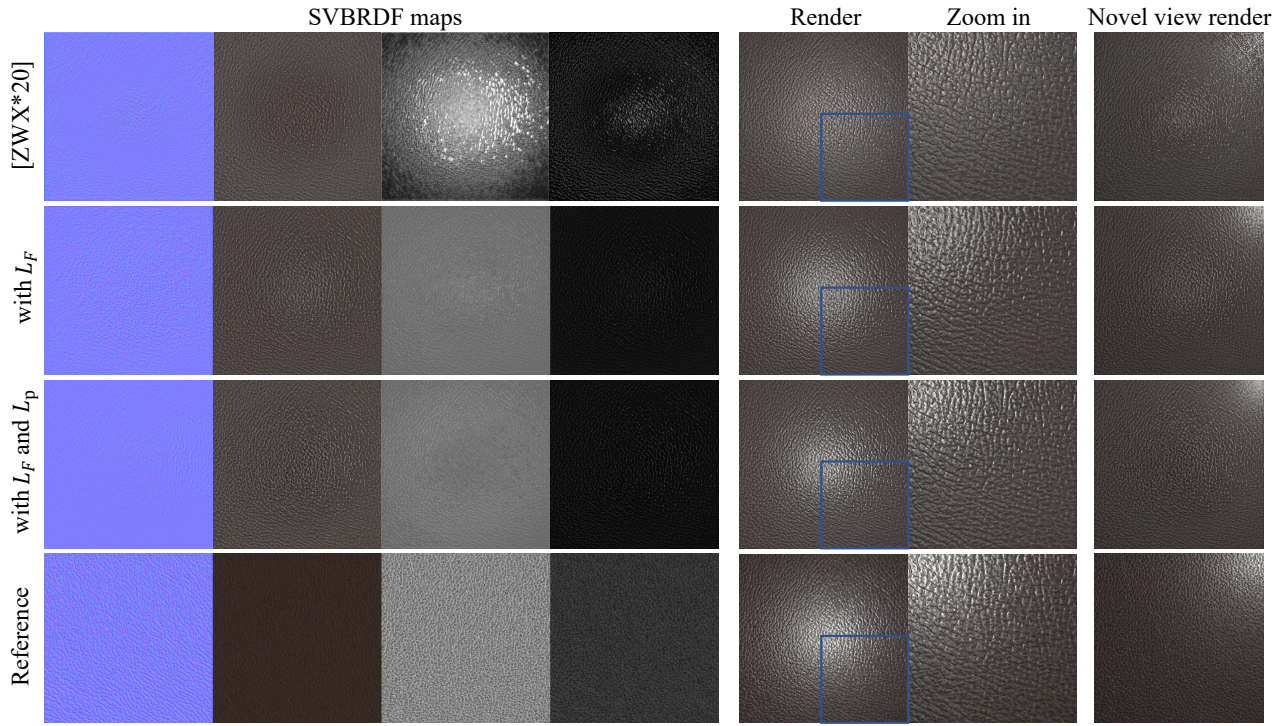}
	\caption{\label{fig:lossleather}%
	Ablation study on several maps to valid the the impacts of Fourier loss and perceptual loss in our SVBRDF recovery network. }
\end{figure*}

\begin{figure*}[htbp]
	\centering
	\includegraphics[width=0.9\linewidth]{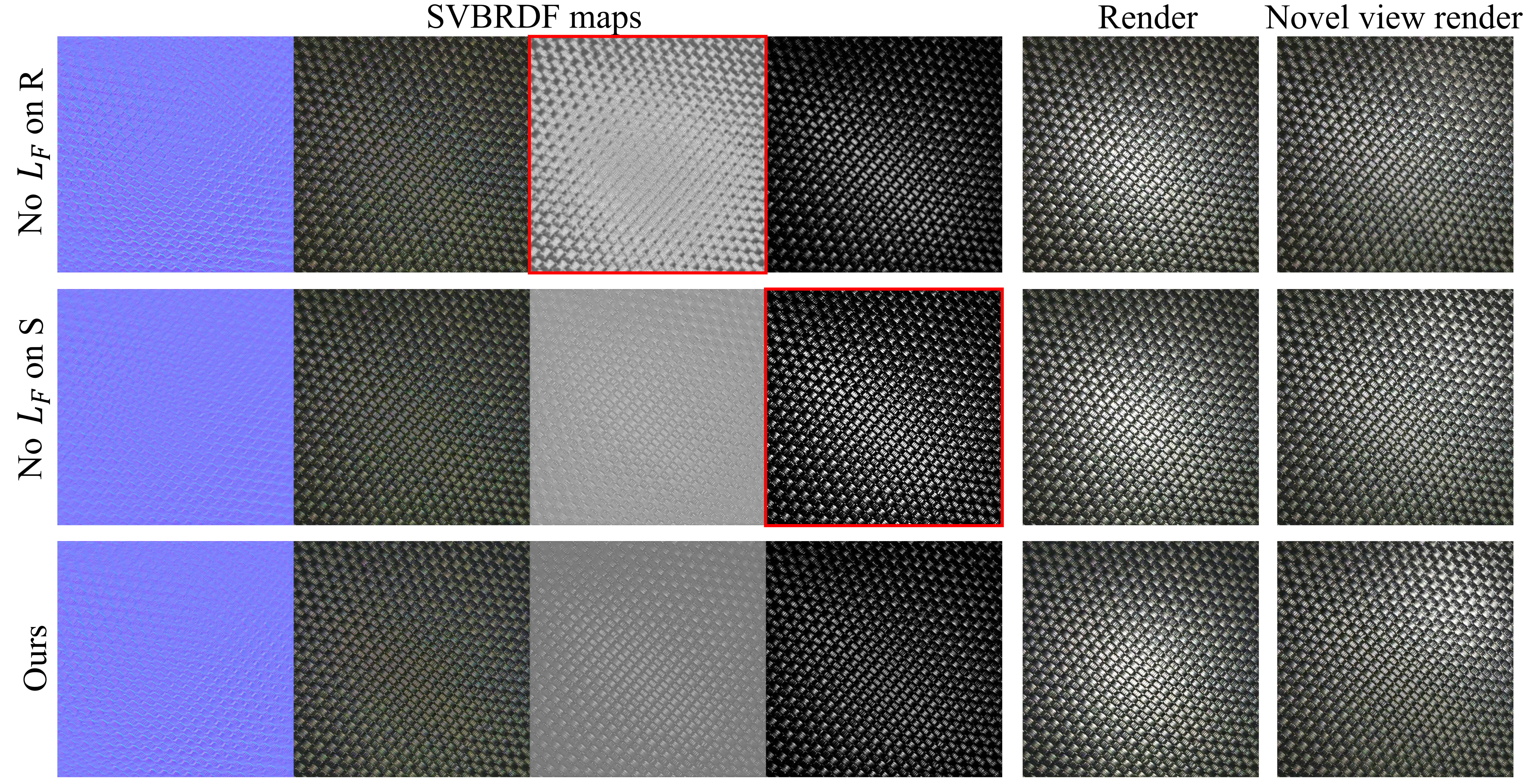}
	\caption{\label{fig:diffFFT}%
	Influence of Fourier loss on different maps. Without Fourier loss on roughness map (R) or specular map (S), the bright spot still exist.}
\end{figure*}

 \begin{figure*}[htbp]
	\centering
	\includegraphics[width=1\linewidth]{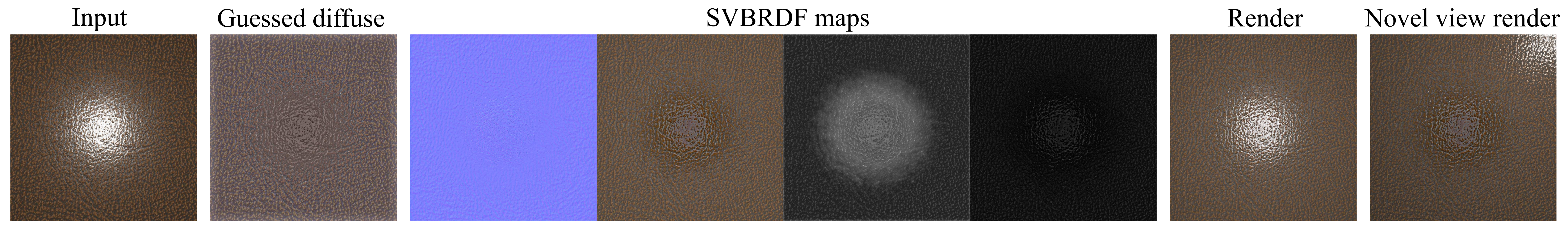}
	\caption{\label{fig:limitation}%
		Failure case. Images with sharp contrast may not be well treated since our method failed to get an stationary ``guessed'' diffuse map.
	}	
\end{figure*}

\begin{figure}[htbp]
	\centering
	\includegraphics[width=1.0\linewidth]{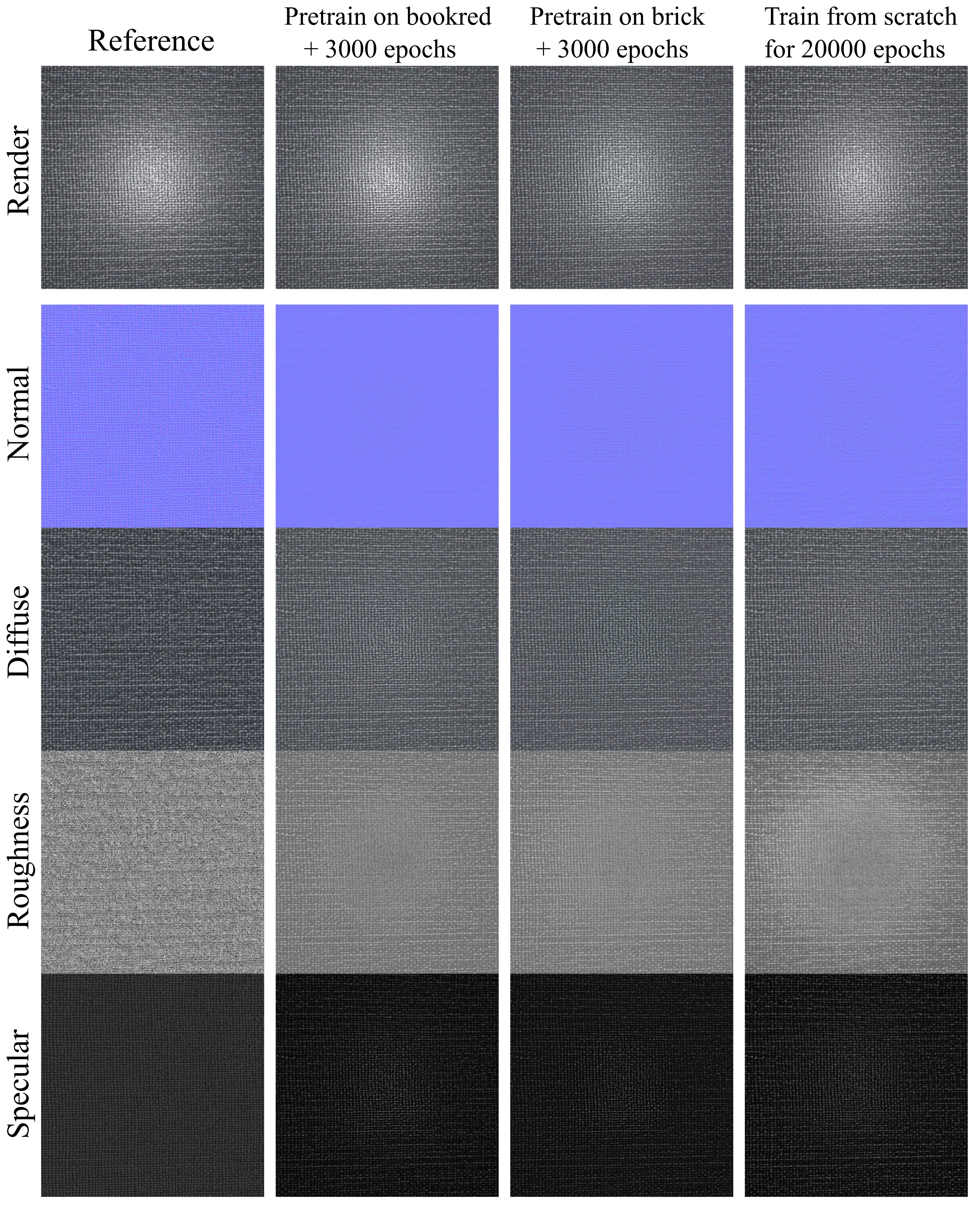}
	\caption{\label{fig:diffmodel}%
	Comparison on different pretrained models and different training strategies. There is little difference in the three results. }
\end{figure}

We first compare the results of our network with Gao et al.~\cite{gao2019deep}, Guo et al.~\cite{Guo2020MaterialGANRC}, Guo et al.~\cite{Guo2021HighlightawareTN} and Zhao et al.~\cite{Zhao2020JointSR} on both synthetic images and captured photos (Sec.~\ref{sec:recovresult}). Then we show the influence of different loss term by a loss ablation experiment (Sec.~\ref{sec:ablationStudy}) and the effects of our two-stage training strategy (Sec.~\ref{sec:twoStageTraining}).

\subsection{Comparison with previous works}
\label{sec:recovresult}

We ran our experiments on images with strong highlights to show the effectiveness of our approach. The input real photos and reference maps are from the two-shot dataset~\cite{aittala2015two} and a free material website$\footnotemark$\footnotetext[1]{https://texturehaven.com}. For the two-shot dataset, we cropped the $3264 \times 2448$ SVBRDF maps to $1600 \times 1600$ and resized them to $1024 \times 1024$ resolution. We render the input synthetic images using the Cook-Torrance reflectance model~\cite{cook1982reflectance} with a point light and camera right above the center of the plane, consistent with our re-rendering process in the network.

\textbf{Comparison on synthetic images.} 
In Figure~\ref{fig:synthetic-book} and Figure~\ref{fig:synthetic-brick}, we compare our results on synthetic images with Gao et al.~\cite{gao2019deep}, Guo et al.~\cite{Guo2020MaterialGANRC}, Guo et al.~\cite{Guo2021HighlightawareTN} and Zhao et al.~\cite{Zhao2020JointSR}. For Gao et al.~\cite{gao2019deep}, we use the network of Deschaintre et al.~\cite{deschaintre2018single} as an initialization and set the number of inputs to one for fair comparison. Although \cite{gao2019deep} achieves plausible rendering results through optimization steps, strong artifacts still exist in the recovered SVBRDF maps, leading to poor performance in novel view rendering. Guo et al.~\cite{Guo2020MaterialGANRC} also produce unpleasant SVBRDF maps and novel view rendering. Zhao et al.~\cite{Zhao2020JointSR} preserve some details in SVBRDF maps but still suffer from highlights regions, especially in roughness map and specular map. Our method recovers the stationarity in SVBRDF maps, which are comparable to the reference maps, thus lead to more plausible appearance in novel view rendering. More results are shown in the supplemental materials.

\textbf{Comparison on captured photos.} 
In Figures~\ref{fig:photo-leather} and \ref{fig:photo-plastic}, we validate our method on three captured photos, comparing to Gao et al.~\cite{gao2019deep}, Guo et al.~\cite{Guo2020MaterialGANRC}, Guo et al.~\cite{Guo2021HighlightawareTN} and Zhao et al.~\cite{Zhao2020JointSR}. The photos are from the two-shot dataset~\cite{aittala2015two}, cropped so that the brightest part of the image is at the center. There are no reference SVBRDFs for the captured photos. As shown in Figure~\ref{fig:photo-leather}, Gao et al.~\cite{gao2019deep} and Guo et al.~\cite{Guo2021HighlightawareTN} produce ``polluted'' SVBRDF maps that are highly affected by highlights regions, while Guo et al.~\cite{Guo2020MaterialGANRC} produce blurred results that lack detailed structure in SVBRDF maps. Zhao et al.~\cite{Zhao2020JointSR} produce plausible diffuse maps, but still suffer from highlights regions in the other maps. All the previous works fail to handle the ambiguity in reflectance and illumination, yielding unpleasant novel view rendering results. Our method produces more stationary SVBRDF maps and more plausible rendering results under novel views. As shown in Figure~\ref{fig:photo-plastic}, our method recovers detailed variations in normal map and suppresses the bright spots in other three maps. Thus, our method better handles the ambiguity in reflectance and illumination, producing more reasonable rendering results under novel view.

\subsection{Ablation study}
\label{sec:ablationStudy}

There are two important components in our SVBRDF recovery network: Fourier loss and perceptual loss. We ran an ablation study to validate the impacts of these components in Figure~\ref{fig:lossleather}. We compare Zhao et al.~\cite{Zhao2020JointSR}, our model (with Fourier loss $\mathcal{L}_{F}$ only), our model (with both Fourier loss $\mathcal{L}_{F}$ and perceptual loss $\mathcal{L}_{P}$) and the reference. 

Zhao et al.~\cite{Zhao2020JointSR} (first row) suffer from bright spots in the roughness and specular maps, and over-flat normal map, leading to obvious difference from the reference SVBRDF maps, where the material properties are stationary. By introducing the Fourier loss,  the predicted maps (second row) become more uniform and have less artifacts at the center. The novel view rendering results have more pronounced variations in illumination due to the more uniform SVBRDF maps, thus produce more plausible appearance compared to Zhao et al.~\cite{Zhao2020JointSR}. However, we found that the texture variation is blurred in the highlights region. Further introducing perceptual loss (third row) solves this issue, via measuring the semantic similarity between the input image and re-rendering result. The joint loss function with both Fourier loss and perceptual loss produces the best results. The bright spots in the SVBRDF maps have been removed, making them decoupled from the illumination in the input images, thus leading to more plausible re-rendering results under different viewing/lighting directions. More results are shown in the supplemental materials.

We also tried using the Fourier loss on only some of the SVBRDF maps, such as the roughness or specular albedos. The results of this study are shown  in Figure~\ref{fig:diffFFT}. The maps that were computed without the Fourier loss are highly affected by the bright spot, while the other maps are not. We used the Fourier loss on all 4 maps to  ensure the stationarity in all SVBRDF maps.

\subsection{Validation of the two-stage training strategy}
\label{sec:twoStageTraining}
In Figure~\ref{fig:diffmodel}, we compare the SVBRDF maps recovered with two-stage training strategy and one-stage training strategy (without pretraining). For the two-stage training strategy, we show the results recovered from two pretrained models which are trained with different images (book-red and brick). For one-stage training strategy (without pretraining), the training is performed on the input image from scratch, with 20,000 iterations. By comparison, we find that the difference between one-stage and two-stage training strategies is subtle, and the difference with different pretrained models are also not obvious. Thus, we believe that our two-stage training strategy greatly reduces the training time, without any quality degradation and the pretrained model can be trained on arbitrary single image under our assumption.

\subsection{Limitations}

We have identified three main limitations for our
method. First, we assume that all maps have the same frequency in their patterns, and that we can use the guessed diffuse map as a guide for learning in the other maps. If the SVBRDF maps have different frequency, our Fourier loss will provide the wrong guidance. Second, our method does not work well for input images with sharp contrast, as shown in Figure~\ref{fig:limitation}. The highlights are so strong that the texture information in this region is almost obscured, which prevent us from getting a plausible guessed diffuse map. Third, although the training time has been reduced significantly, compared to the previous work, each image still needs 30 minutes for training. Further reducing the training time will be a valuable work, and we leave it for the future work.


\section{Conclusion}
\label{sec:conclusion}
In this paper, we improve the unsupervised SVBRDF recovery generative adversarial network, by introducing two new loss functions: a Fourier loss function and a perceptual loss function to enforce the stationarity of SVBRDFs, yielding better quality for the SVBRDF maps, especially for input images with intense highlights. Then, we propose a two-stage training strategy to reduce the training time with 8$\times$ speedup. In the end, our method is able to generate high-quality SVBRDFs and produce more plausible rendering results, compared with the state-of-the-art methods.


\bibliographystyle{eg-alpha-doi} 
\bibliography{egbibsample}       


\end{document}